\documentclass{article}

\usepackage{times}
\usepackage{amsmath,amsfonts,amssymb,amsthm,enumitem}
\usepackage{graphicx}
\usepackage{xcolor}
\usepackage{natbib}
\usepackage{mathpazo}
\usepackage{algorithm}
\usepackage{algorithmic}
\usepackage{lipsum}
\usepackage{comment}
\usepackage{hyperref}
\usepackage{tabularx}
\usepackage{array}
\usepackage{todonotes}
\usepackage{subfig}
\usepackage{multirow}

\usepackage[accepted]{icml2019}

\DeclareMathOperator{\E}{\mathbb{E}}
\DeclareMathOperator{\Real}{\mathbb{R}}
\DeclareMathOperator{\Ncal}{\mathcal{N}}
\DeclareMathOperator{\Wcal}{\mathcal{W}}
\DeclareMathOperator{\Xcal}{\mathcal{X}}
\DeclareMathOperator{\Ycal}{\mathcal{Y}}
\DeclareMathOperator{\Dcal}{\mathcal{D}}
\DeclareMathOperator{\Ccal}{\mathcal{C}}

\DeclareMathOperator{\Dir}{\mathcal{D}\textit{ir}}

\DeclareMathOperator{\x}{\mathbf{x}}

\DeclareMathOperator{\y}{\mathbf{y}}

\icmltitlerunning{Output-Constrained Bayesian Neural Networks}

\begin{document}

\twocolumn[
\icmltitle{Output-Constrained Bayesian Neural Networks}

\icmlsetsymbol{equal}{*}

\begin{icmlauthorlist}
\icmlauthor{Wanqian Yang}{equal,harvard}
\icmlauthor{Lars Lorch}{equal,harvard}
\icmlauthor{Moritz A. Graule}{equal,harvard}
\icmlauthor{Srivatsan Srinivasan}{harvard}
\icmlauthor{Anirudh Suresh}{harvard}
\icmlauthor{Jiayu Yao}{harvard}
\icmlauthor{Melanie F. Pradier}{harvard}
\icmlauthor{Finale Doshi-Velez}{harvard}
\end{icmlauthorlist}

\icmlaffiliation{harvard}{Harvard University}

\icmlcorrespondingauthor{Wanqian Yang}{yangw@college.harvard.edu}
\icmlcorrespondingauthor{Lars Lorch}{lars.lorch@gmail.com}

\icmlkeywords{Machine Learning, ICML}

\vskip 0.3in
]

\printAffiliationsAndNotice{\icmlEqualContribution} 

\begin{abstract}
Bayesian neural network (BNN) priors are defined in parameter space, making it hard to encode prior knowledge expressed in function space.
We formulate a prior that incorporates functional constraints about what the output can or cannot be in regions of the input space.
Output-Constrained BNNs (OC-BNN) represent an interpretable approach of enforcing a range of constraints, fully consistent with the Bayesian framework and amenable to black-box inference.
We demonstrate how OC-BNNs improve model robustness and prevent the prediction of infeasible outputs in two real-world applications of healthcare and robotics.
\end{abstract}

\vspace{-0.5cm}


\section{Introduction}
\label{sec:introduction}

BNNs combine powerful function approximators with the ability to model uncertainty, making them useful in domains where (i) training data is expensive or limited, or (ii) inaccurate predictions are prohibitively costly and decision-making must be informed by our level of confidence \cite{mackay, neal}.
Domain experts often have prior knowledge about the modeled function and the ability to encode such information on top of training data can thus improve performance.
However, BNNs define prior distributions over parameters, whose high dimensionality and lack of interpretability make the incorporation of functional beliefs close to impossible.
We present an interpretable approach for incorporating prior functional information into BNNs in the form of constraints, while staying consistent with the Bayesian framework.
We then apply our method to two domains where the ability to encode such constraints is crucial:
\textbf{(i)} prediction of clinical actions in health care, where constraints prevent unsafe actions for certain physiological inputs, and
\textbf{(ii)} human motion prediction, where joint positions are constrained by anatomically feasible ranges.
Our contributions are:
\textbf{(a)} we introduce \textit{constraint priors}, capable of incorporating both \emph{negative} constraints (where the function cannot be) and \emph{positive} constraints (where the function should be), applicable with any black-box inference algorithm normally used with BNNs,
and \textbf{(b)} we demonstrate the application of constraint priors with a variety of suitable inference methods on toy problems as well as two large and high-dimensional real-world data sets.


\section{Related Work}
\label{sec:related}
Most closely related to our work, \cite{lorenzi} considered function-space equality and inequality constraints of deep probabilistic models. However, they focused on deep Gaussian processes (DGPs) rather than BNNs, and on low-dimensional data from simulated ODE systems, whereas we consider high-dimensional real-world settings.  They also do not consider classification settings.

\cite{ncp} specify a Gaussian function prior with the goal of preventing overconfident BNN predictions out-of-distribution.
In contrast, we use "positive constraints" to guide the function where it should be.
Also related are functional BNNs by \cite{fbnn}, where variational inference is performed in function-space using a stochastic process model. Their view is more general---and accordingly, more complex to optimize---while we focus on constraints in specific regions of the input-output space.

\section{Background}
\label{sec:background}
\setlength{\abovedisplayskip}{5pt}
\setlength{\belowdisplayskip}{5pt}
A conventional BNN, operating in the function (or input-output) space $\Xcal \times \Ycal$, typically has a prior over parameters $p(\Wcal)$, where $\Wcal$ are the neural network weights and biases.
Given data $\Dcal = \{ \x_n, \y_n \}_{n=1}^N$, we perform inference to obtain the posterior $p(\Wcal | \Dcal) \propto p(\Wcal)p(\Dcal|\Wcal)$. The posterior predictive for the output $\y'$ for some new input $\x'$ is obtained by integrating over the posterior distribution of $\Wcal$:
\begin{align}\label{eq:posterior_predictive}
    p(\y' | \x', \Dcal) = \int_{\Wcal} p(\y' | \x', \Wcal) p(\Wcal | \Dcal) d\Wcal
\end{align}

The space of $\Wcal$ is high-dimensional and the relationship between the weights and the function is non-intuitive.  As such, the prior $p(\Wcal)$ is often trivially chosen as an isotropic Gaussian: 
\begin{equation}
    p(\Wcal) = \prod_i \Ncal(\Wcal_i; 0, \sigma_p^2)
    \label{eq:weight_prior_term}
\end{equation}

\section{Output-Constrained BNNs}
\label{sec:theory}

We consider two kinds of ``expert knowledge'': \textit{positive} constraints define regions where a function \textit{should} be, and \textit{negative} constraints define regions where a function \textit{cannot} be.
This delineation is not arbitrary --- the level of prior knowledge (strongly vs. weakly informative) and the task (regression or classification) may suggest the use of different prior constraints.

\textbf{Defining constrained regions} \quad Formally, a \textit{positive} constrained region $\mathcal{C}^{+}$ is a set of input-output tuples $(\x, \y)$ defining where outputs given certain inputs should be.
Conversely, a \textit{negative} constrained region $\mathcal{C}^{-}$ is a set of tuples $(\x, \y)$ defining where outputs given certain inputs cannot be.
We will use $\Ccal$ when describing properties of constrained regions of both kinds and denote $\Ccal_{\x}$ for all $\x$ in $\Ccal$ and $\Ccal_{\y}$ for all $\y$ in $\Ccal$.
Given this formulation, it is our goal to enforce
\begin{align}\label{eq:constraint_prior_goal}
\begin{split}
\hspace*{-1.5mm}\int_{\Wcal} p(\y'\notin \Ccal^{+}_{\y} | \x' \in \Ccal^{+}_{\x}, \Wcal) p(\Wcal | \Ccal^{+}, \Dcal) d\Wcal &\approx 0\\
\hspace*{-1.5mm}\int_{\Wcal} p(\y'\in \Ccal^{-}_{\y} | \x' \in \Ccal^{-}_{\x}, \Wcal) p(\Wcal | \Ccal^{-}, \Dcal) d\Wcal &\approx 0\\
\end{split}
\end{align}
Note that (\ref{eq:constraint_prior_goal}) is simply the posterior predictive distribution conditioned on $\Ccal$.
The generality of this approach allows for the incorporation of very complicated yet interpretable constraints \textit{a priori}, such as for example arbitrary equality, inequality and logical (if-then and either-or) constraints.

\textbf{Constraint prior} \quad We connect the weight space of the BNN with constraints through the distribution:
\begin{align}
    g(W|\Ccal) = g(\phi(\Ccal_{\x}; \Wcal); \Ccal_{\y}, \theta)
    \label{eq:g_term}
\end{align}
where $\phi(\x; \Wcal)$ is the BNN forward pass and $\theta$ is the set of tuneable hyperparameters of $g$. Accordingly, a \textit{constraint prior} $p_{\Ccal}(\Wcal)$ can then be constructed as:
\begin{align}\label{eq:constraint_prior_general}
   p_{\Ccal} (\Wcal) := p(\Wcal)\;g(\Wcal|\Ccal),
\end{align}
achieving the goal of expressing prior function knowledge in weight space while retaining the weight-space prior $p(\Wcal)$.
Intuitively, $g(\Wcal|\Ccal)$ measures the BNN's adherence to the constrained region.
It remains to describe how $g$ is defined. For positive constraints $\Ccal^{+}$, $g$ measures how close $\phi(\Ccal^{+}_{\x}; \Wcal)$ lies to $\Ccal_{\y}$, for which natural choices of distributions exist for both regression and classification.
For negative constraints $\Ccal^{-}$, we define $g$ as the expected violation of $\Ccal^{-}_{\y}$ given $\phi(\Ccal^{-}_{\x}; \Wcal)$ using a classifier function.
Complete definitions of $g$ for positive and negative priors are provided in Appendix~\ref{app:constraints}; details on inference procedures are provided in Appendix~\ref{app:inference}.

\section{Demonstrations on Synthetic Data}
\label{sec:toyexperiments}

This section provides proof of concepts of OC-BNNs using 2-dimensional synthetic examples.
Refer to Appendix \ref{app:experimental} for experimental details and Appendix \ref{app:additional} for additional results. For regression, the posteriors are visualized in black/gray for baseline BNNs, and blue for OC-BNNs. Negative constrained regions are red; positive (Gaussian) constraints are green. For the classification example, the three classes are color-coded red, green and blue.

\begin{figure}[H]
\vspace*{0.5em}
\centering
\includegraphics[width=0.48\columnwidth]{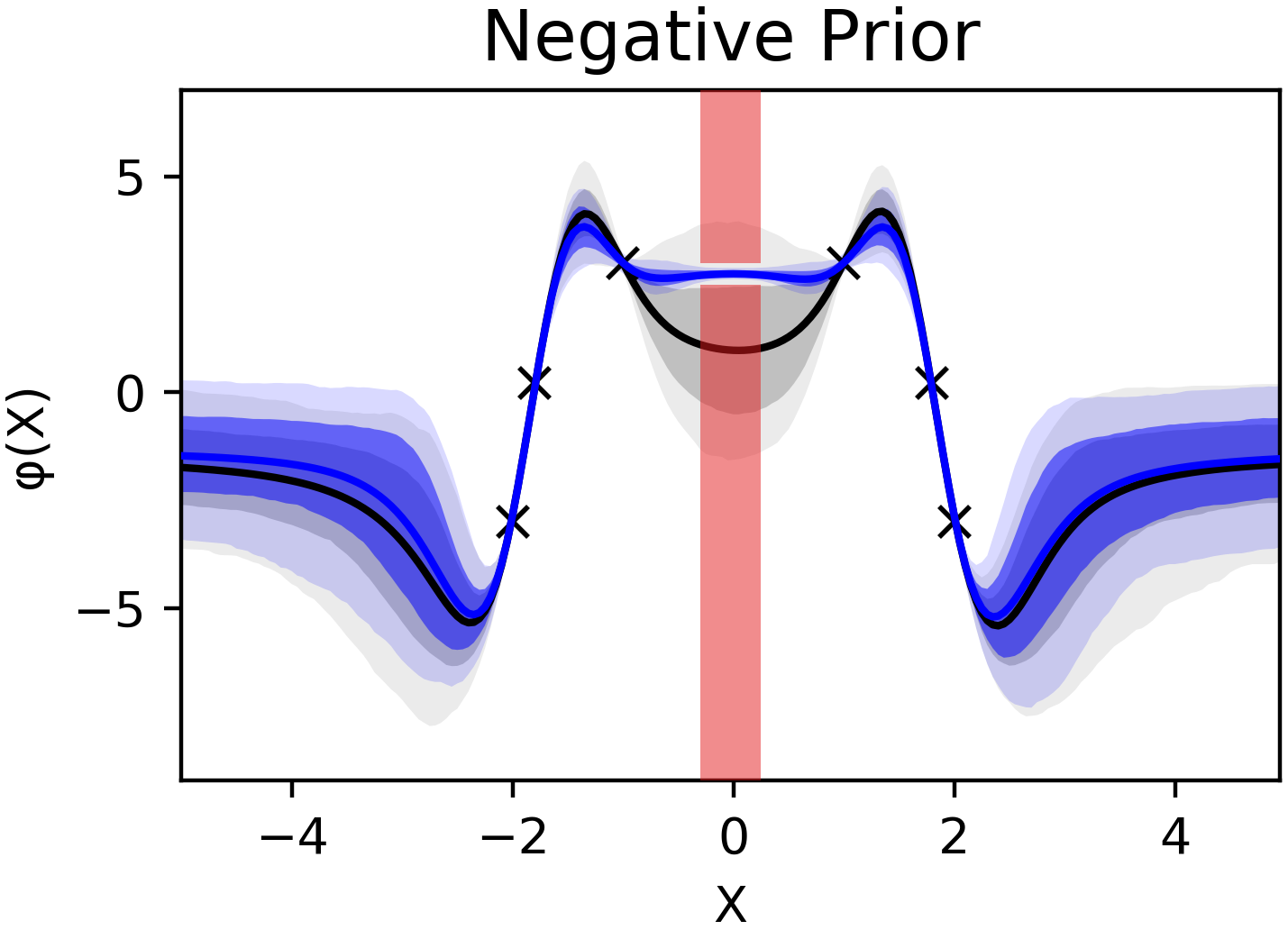}
\includegraphics[width=0.50\columnwidth]{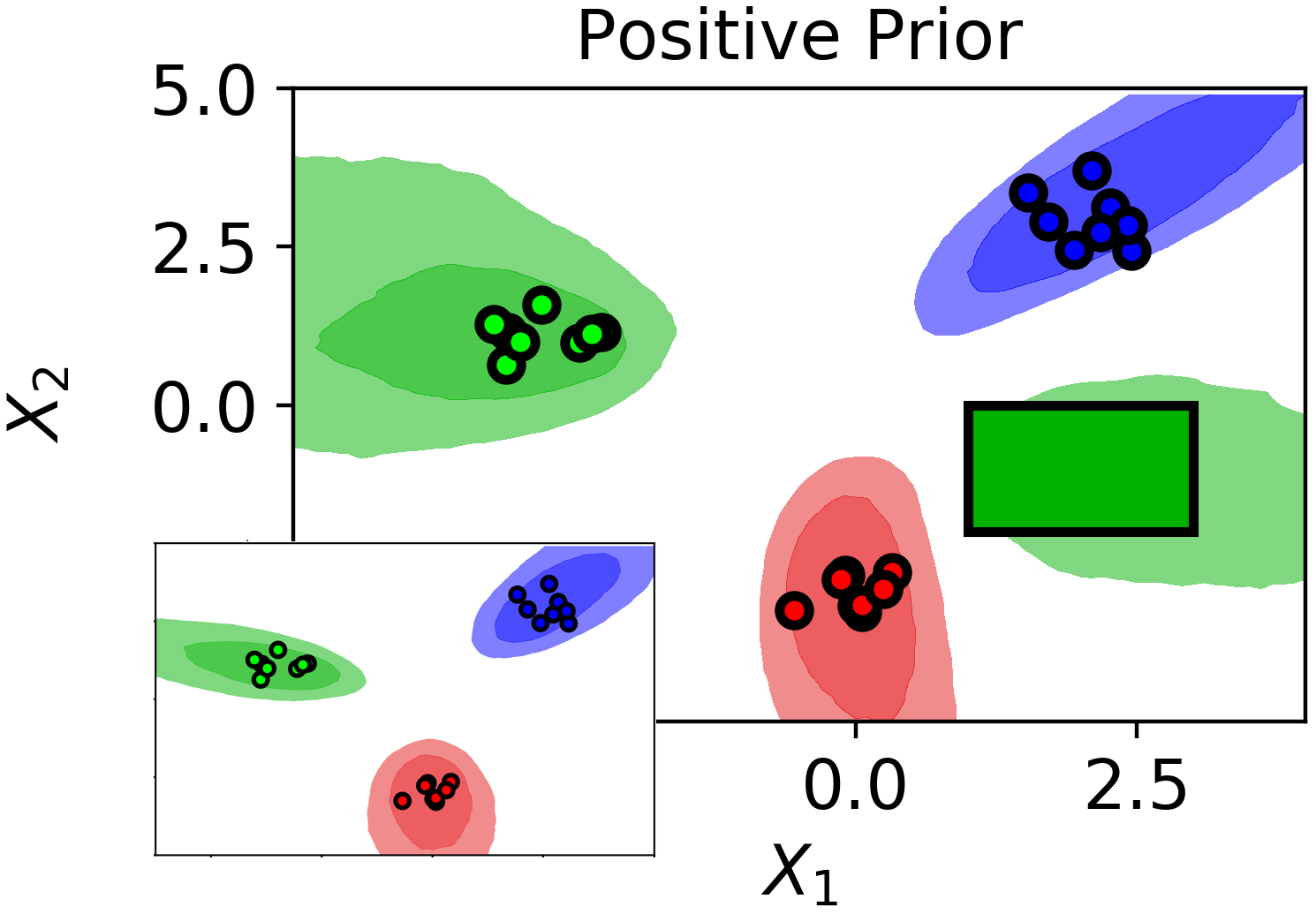}
\vspace*{-1.0em}
\caption{
On both tasks, OC-BNNs reduce uncertainty in constrained regions while fitting data well. \textbf{(left)} 1D regression. The constraint is composed of two negative regions (red) separated by a small gap. Uncertainty of OC-BNNs (blue) drops sharply in the constrained region compared to the baseline (gray). \textbf{(right)} 2D classification with three classes. Constrained region enforces the prediction of green class in the green rectangle (baseline depicted in inset).}
\label{fig:ex14}
\end{figure}

\textbf{OC-BNNs model uncertainty in a manner that respects constrained regions while explaining training data.}
Figure~\ref{fig:ex14} demonstrates this for both the regression and classification setting. Correct predictions are maintained
with similar uncertainty levels as the baseline while constraints are correctly enforced with uncertainty levels changing to reflect that. These examples demonstrate how OC-BNNs enforce constraints without sacrificing predictive accuracy.

\begin{figure}[H]
\vspace*{0.5em}
\includegraphics[width=0.49\columnwidth]{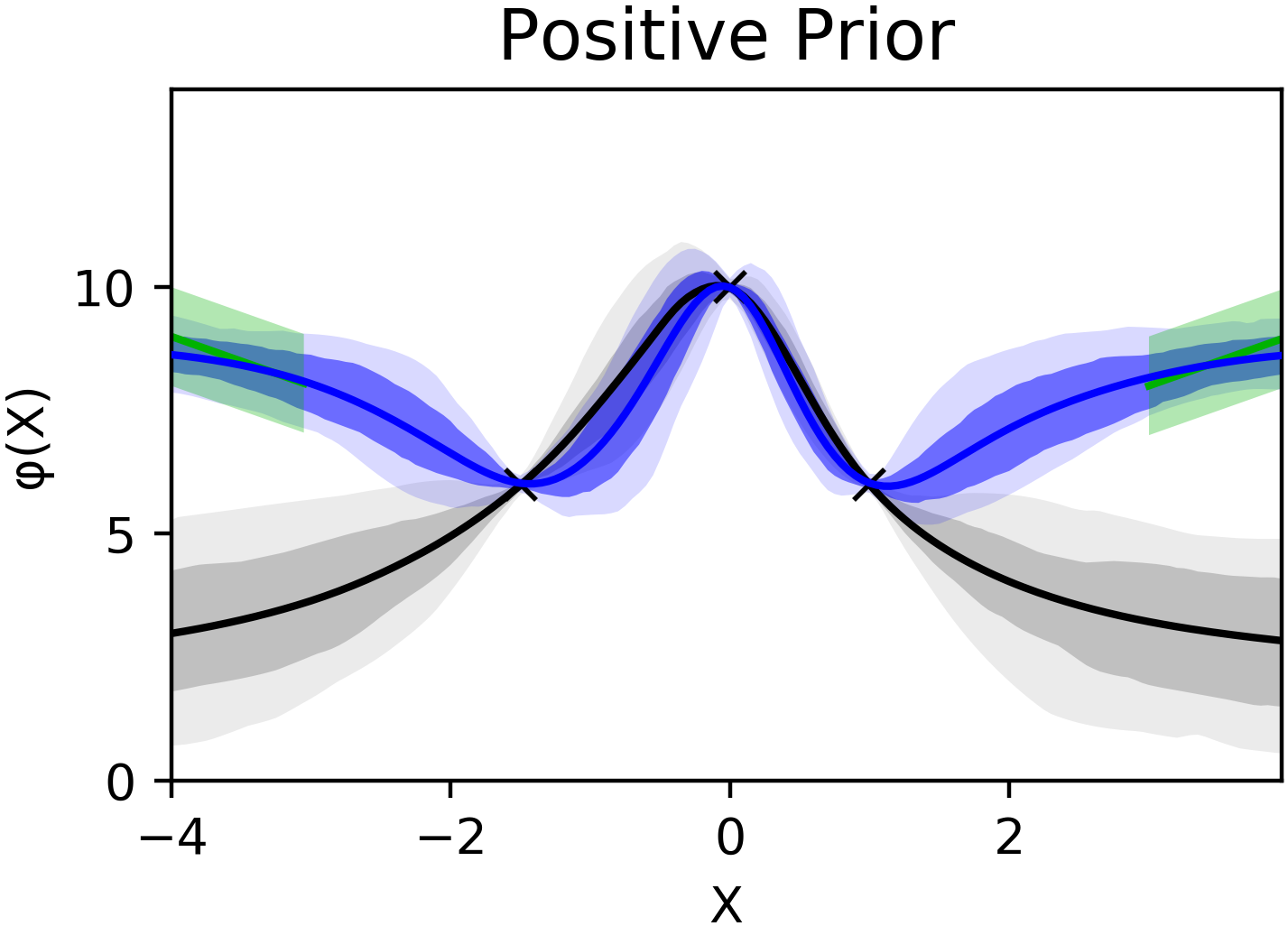}
\includegraphics[width=0.49\columnwidth]{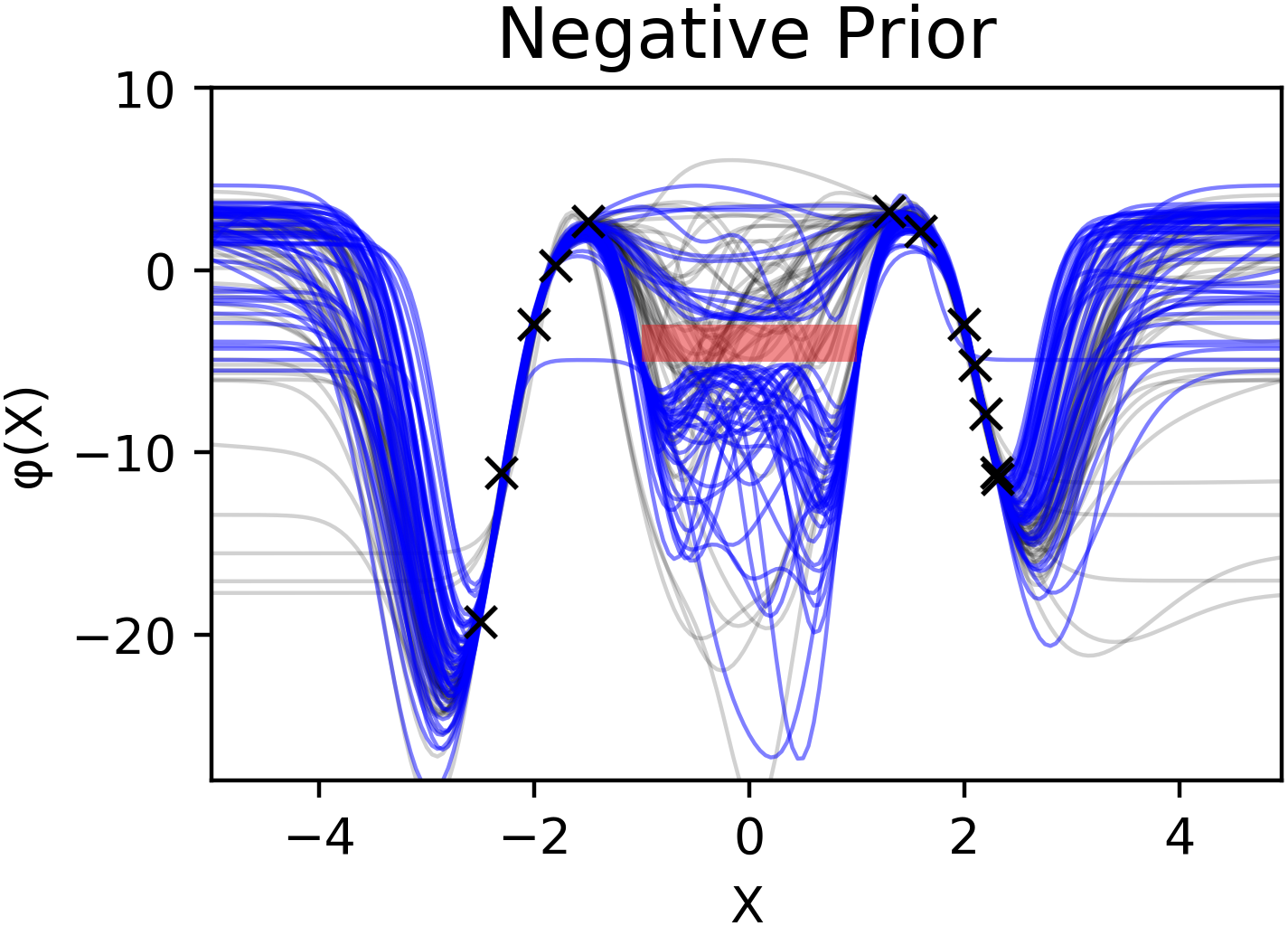}
\vspace*{-1.0em}
\caption{OC-BNNs capture important posterior qualities such as correct OOD behavior and multimodality. \textbf{(left)} OC-BNNs (blue) maintain the same in-distribution uncertainty as the baseline (gray) while adhering to OOD positive constraints (green) on either side of the plot. \textbf{(right)} OC-BNNs (blue) posterior samples go both below and above the negative constraint box (red).}
\label{fig:ex35}
\end{figure}

\textbf{OC-BNNs encourage correct out-of-distribution behavior.} Figure~\ref{fig:ex35} (left) depicts sparse data, along with out-of-distribution positive constrained regions. The posterior predictive \textit{in-distribution} closely mimics the baseline, while the posterior \textit{out-of-distribution} (OOD) learns to avoid the constrained region. This demonstrates that OC-BNNs function well away from the data, which is important because we typically want to enforce functional constraints when there is a lack of observed training data for the model to learn from.

\textbf{OC-BNNs can capture posterior multimodality.} While negative constraints $\Ccal^{-}$ do not explicitly define multimodal posterior predictives, a bounded constrained region does imply that the posterior predictive might have probability mass on either side of the bounded region (i.e. for all $d$ dimensions of $\Ycal = \Real^d$). Figure~\ref{fig:ex35} (right), demonstrates that we capture challenging posterior predictives.

\section{Applications}
\label{sec:applications}

\subsection{Clinical action prediction}
\label{sec:application_heydoc}

MIMIC-III \cite{mimic} is a benchmark database containing time series data of various physiological measurements and clinical actions prescribed belonging to $>40,000$ intensive care patients who stayed at the Beth Israel Deaconess Medical Center between 2001 and 2012.

\textbf{Problem Formulation} \quad From the raw time-series data, we construct a balanced dataset for a time-independent classification task of hypotension management. There are 9 features representing various physiological states, such as mean blood pressure and lactate levels.  The goal is to predict if clinical action (either vasopressor or IV fluid) should be taken.

\textbf{Constraints} \quad The constraint imposed is that for \texttt{mean blood pressure} less than 65 units, some action should be taken, which is physiologically realistic. We apply the positive (Dirichlet) constraint prior (Appendix~\ref{app:constraints}), as well as the weights-only prior baseline. In the given data, some training points fall within the constrained region. We train our model both with and without artificially filtering out all points within the positive constrained region.

\textbf{OC-BNNs maintain classification accuracy while reducing physiologically infeasible constraint violations.} \quad Table~\ref{results:icu} displays experimental results, with statistics computed from the posterior mean. In addition to standard accuracy (ACC) and F1 score, we measure the violation fraction (VIOL), which is the fraction of predictions on held-out points that violate the constraints. The results show that OC-BNNs match standard BNNs on all predictive accuracy metrics, with significantly lower violation of the constrained region for the case where points originally in the constrained region are filtered out.

\begin{table}[]
\begin{tabular}{cl|r|r|r|r|}
\cline{3-6}
\multicolumn{1}{l}{} &  & \multicolumn{2}{c|}{filtered} & \multicolumn{2}{c|}{unfiltered} \\ \cline{3-6}
\multicolumn{1}{l}{} &  & \multicolumn{1}{c|}{\textbf{BNN}} & \multicolumn{1}{c|}{\hspace*{-1mm}\textbf{OC-BNN}\hspace*{-1mm}} & \multicolumn{1}{c|}{\textbf{BNN}} & \multicolumn{1}{c|}{\hspace*{-1mm}\textbf{OC-BNN}\hspace*{-1mm}} \\ \hline
\multicolumn{1}{|c|}{\multirow{3}{*}{\rotatebox[origin=c]{90}{\textbf{Train}}}} & ACC & 0.745 & 0.741 & 0.881 & 0.878 \\ \cline{2-6}
\multicolumn{1}{|c|}{} & F1 & 0.805 & 0.801 & 0.882 & 0.880 \\ \cline{2-6}
\multicolumn{1}{|c|}{} & VIOL & 0.151 & 0.149 & N/A & N/A \\ \hline\hline
\multicolumn{1}{|c|}{\multirow{3}{*}{\rotatebox[origin=c]{90}{\textbf{Test}}}} & ACC & 0.660 & 0.665 & 0.647 & 0.649 \\ \cline{2-6}
\multicolumn{1}{|c|}{} & F1 & 0.746 & 0.748 & 0.725 & 0.736 \\ \cline{2-6}
\multicolumn{1}{|c|}{} & VIOL & 0.132 & 0.126 & \textbf{0.117} & \textbf{0.039} \\ \hline
\end{tabular}
\caption{Results for the MIMIC experiments with and without filtering out the points in the constrained region. Accuracy and F1 score remain unchanged when using OC-BNNs. For the experiment with filtration, the violation factor decreases by a factor of 3 when using OC-BNNs.
\vspace*{-1.5em}}
\label{results:icu}
\end{table}

\subsection{Human motion prediction}
\label{sec:applications_human}

We evaluate OC-BNNs on data of humans conducting various motions available at  \cite{kratzer2019data} as described in \cite{kratzer2018towards}.
This data contains human upper body poses across many reaching tasks at a frame rate of 120Hz. The poses are provided in the form of upper body joint angles.
\textbf{Problem formulation} \quad
Given a subset of trajectories in \cite{kratzer2019data}, our goal is to predict joint angles 20 frames in the future from angles at the current time frame and the numerically computed joint velocities and accelerations. In the following, we limit ourselves to abduction and flexion (further denoted as Y- and Z-rotation to match the nomenclature in the original data \cite{kratzer2019data}) of the left and right shoulder during right-handed reaching motions.
The joint angles in the test data were perturbed with normally distributed noise ($\mu$$=$$0$, $\sigma$$=$$2$ degrees) to simulate a scenario in which a human motion prediction model is trained on data recorded in a high-end motion capture lab, and then used to predict  motion from data obtained by noisy wearable sensors.
\textbf{Constraints} Several anatomical feasibility or functional range constraints for each of the joint angles could be applied, e.g. as described in \cite{namdari2012defining}.
We derived constraints on the joint limits from the reaching motions provided in \cite{kratzer2019data} as the empirically observed extrema
across all motions, which is modeled using the negative constraint prior.
\textbf{OC-BNNs prevent infeasible predictions.}
We compare a BNN and OC-BNN using the negative prior and the empirical bounds on joint angles.
Both models are compared in
(i) RMSE using the posterior predictive mean (RMSE) [$1$$\cdot$$10^3$],
(ii) held-out data log likelihood of  $\Ncal(\mu_{pp}, \sigma_{pp}^2)$ with posterior predictive mean $\mu_{pp}$ and variance $\sigma_{pp}$ (HO-LL), and
(iii) posterior predictive violation defined as the percentage of probability mass in an infeasible constrained region (PP-VIOL) [\%],
each evaluated at all target points.

\begin{figure}[h]
\vspace*{+1.0em}
\centering
\subfloat{\includegraphics[width=0.95\columnwidth]{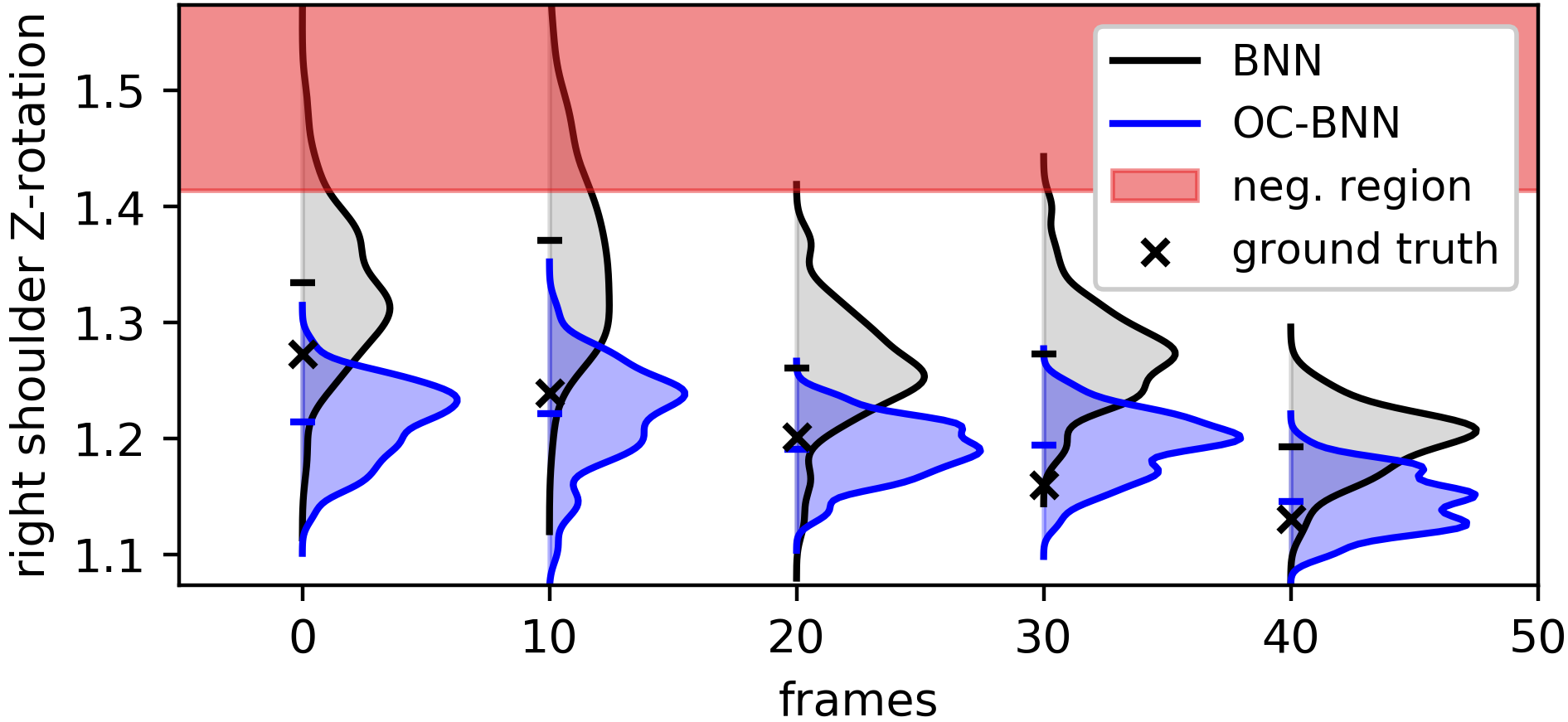}}
\vspace*{-1.0em}
\caption{%
Consecutive predictions of right Z-rotation during an exemplary test trajectory with 10 frame gaps.
The posterior predictive of BNN (black) and OC-BNN (blue) given the input state 20 frames earlier are plotted along the y-axis.
While both BNN and OC-BNN do not perfectly generalize to the test set, the expert knowledge of valid joint positions enforces feasible and thus more robust predictions.}
\label{fig:robots}
\end{figure}

These metrics are summarized in Table \ref{results:human_motion}.
We find that OC-BNNs reduce the possibility of making an infeasible prediction to less than 0.001\%, substantially improving on BNNs.
Figure \ref{fig:robots} shows exemplary motion predictions obtained with both BNN and OC-BNN for five consecutive points in a test trajectory.
\vspace{4pt}
\begin{table}[H]
\centering
\begin{tabular}{ll|rr|}
\cline{3-4} &  & \multicolumn{1}{c|}{\textbf{BNN}} & \multicolumn{1}{c|}{\textbf{OC-BNN}}  \\ \hline
\multicolumn{1}{|l|}{\multirow{3}{*}{\rotatebox[origin=c]{90}{\textbf{Train}}}} & RMSE & \multicolumn{1}{r|}{0.929} &  1.252 \\ \cline{2-4}
\multicolumn{1}{|l|}{} & HO-LL & \multicolumn{1}{r|}{1718.409} & 1342.602 \\ \cline{2-4}
\multicolumn{1}{|l|}{} & PP-VIOL & \multicolumn{1}{r|}{\textbf{0.046}} &\textbf{ 0.000} \\ \hline\hline
\multicolumn{1}{|l|}{\multirow{3}{*}{\rotatebox[origin=c]{90}{\textbf{Test}}}} & RMSE & \multicolumn{1}{r|}{7.320} & 12.127 \\  \cline{2-4}
\multicolumn{1}{|l|}{} & HO-LL & \multicolumn{1}{r|}{101.129} & -683.697 \\ \cline{2-4}
\multicolumn{1}{|l|}{} & PP-VIOL & \multicolumn{1}{r|}{\textbf{18.447}} &\textbf{ 0.000 }\\ \hline
\end{tabular}
    \caption{Results for human motion prediction. While predictive performance and held-out log likelihood are similar, OC-BNNs (negative prior) reduce the chance of predicting an infeasible position to 0.0 \% while BNNs make infeasible predictions in 18.4\% of cases.}
    \label{results:human_motion}
\end{table}

\section{Discussion}
\label{sec:discussion}

\textbf{OC-BNNs prevent constraint violation while fitting low- and high-dimensional data.} Our results highlight that incorporating expert knowledge into OC-BNNs helps enforcing feasible and thus more robust predictions. Results for both datasets in Section \ref{sec:applications} demonstrate that constraint violation metrics are reduced significantly, whereas accuracy metrics are nearly unchanged. This affirms the behavior observed in the synthetic examples in Section~\ref{sec:toyexperiments}.

\textbf{Training data in constrained region can outweigh prior effect.} The clinical dataset results show that the presence of data in $\Ccal$ reduces the effect of constraint priors.
This is expected and in accordance with the Bayesian framework, where the likelihood effect will crowd out the prior given enough training data, and also suggests that the practitioner can use OC-BNNs even for situations where the constraints themselves may not be fully satisfied.

\textbf{OC-BNNs can facilitate data imputation.} The fact that OC-BNNs model uncertainty correctly in constrained regions without losing predictive accuracy, even for high-dimensional datasets, show that OC-BNNs can encode imputation in input regions without training data. Rather than directly modifying the training set through imputation, prior beliefs about missing data can instead be formulated as constraints.

\textbf{When to use which prior?} In the regression setting, negative priors are weakly informative whereas positive priors tend to be strongly informative -- one or both of the prior types can be used depending on domain knowledge. While the negative prior formulation does not apply to classification cases, this does not pose a problem as negative and positive constraints are complements in discrete space.

\section{Conclusion and Outlook}
\label{sec:conclusion}
We describe OC-BNNs, a formulation to incorporate expert knowledge into BNNs by prescribing positive and negative (i.e., desired and forbidden) regions, and demonstrate their application to synthetic and real-world data. We show that OC-BNNs generally maintain the desirable properties of regular BNNs while their predictions follow the prescribed constraints. This makes them a promising tool for settings like healthcare, where models trained on sparse data may be augmented with expert knowledge. In addition, OC-BNNs may find applications in safe reinforcement learning, e.g. in tasks where certain actions are known to have catastrophic consequences.

\section*{Acknowledgements}

MG and FDV acknowledge support from AFOSR FA 9550-17-1-0155. LL and WY acknowledge support from the John A. Paulson School of Engineering and Applied Sciences at Harvard University.


\newpage
\appendix

\section{Constraint Priors}
\label{app:constraints}
In this section, we describe the detailed functional forms of our positive and negative constraints and priors for both classification and regression settings, noting aspects important for inference.

\subsection{Positive constraint prior}\label{ssec:positive_prior}
Since $\Ccal^{+}$ describes the set of points that the learned function should model, $g(\phi(\Ccal_{\x}; \Wcal); \Ccal_{\y}, \theta)$ has the straightforward interpretation of measuring how closely $\phi(\Ccal_{\x}; \Wcal)$ lies to $\Ccal_{\y}$. Most common probability distributions as well as (possibly improper) user-defined distributions are amenable, though differentiability may be a condition for certain inference methods. In particular, natural choices of distributions exist for both regression and classification.

\textbf{Regression} \quad In the simplest setting, for which there is a known ground-truth function described perfectly by $\Ccal^{+}$, the Gaussian distribution is a natural choice:
        \begin{align}\label{eq:pp_gaussian}
        g(\Wcal| \Ccal^{+}) = \prod_{\x, \y \sim \pi( \Ccal^{+})} \Ncal(\phi(\x; \Wcal) ; \y, \sigma_{+}^2)
        \end{align}
where $\pi(\Ccal^{+})$ is a sampling distribution for $\Ccal^{+}$, which is necessary for tractability if $\Ccal^{+}$ is large or infinite.
$\pi(\Ccal^{+})$ itself can be user-defined as the domain allows, allowing for flexibility in sampling.
$\sigma_{+}$ is the tuneable standard deviation of the Gaussian, controlling strictness of deviation from $\Ccal^{+}$. More generally, it is possible that there exists multiple $\y \in \Ccal^{+}_{y}$ for some $\x \in \Ccal^{+}_{x}$. This can be expressed using multimodal distributions, for example:
\begin{align}\label{eq:pp_kgaussian}
    g(\Wcal| \Ccal^{+}) = \hspace*{-5mm} \prod_{\x, \{\y\}_k^K \sim \pi( \Ccal^{+})} \sum^K_{k=1} \omega_k \Ncal(\phi(\x; \Wcal) ; \y_k, \sigma_{+}^2)
\end{align}
where $\omega_k$ are the user-defined mixture weights.

\textbf{Classification} \quad $\Ccal^{+}_{\y}$ describes the classes that the BNN is constrained to for the corresponding $\Ccal^{+}_{\x}$. In the discrete setting, the natural distribution is the Dirichlet. For $K$ classes,
\begin{align}\label{eq:pp_kgaussian}
    g(\Wcal| \Ccal^{+}) = \prod_{\x, \y \sim \pi( \Ccal^{+})} \Dir(\phi(\x; \Wcal) ; \alpha)
\end{align}
where $\alpha_k = \begin{cases} 1 &\text{if $y_k = 1$}\\1 - \alpha_\sigma &\text{otherwise}\end{cases}$ for some controllable penalty $\alpha_\sigma$.
\subsection{Negative constraint prior} \label{ssec:negative_prior}
The negative constraint prior enforces the infeasibility of regions in function space and is constructed by placing little prior probability on high expected violation of $\Ccal^{-}$:
\begin{align}\label{eq:negative_prior}
    \hspace*{-2mm}g(\Wcal| {\Ccal^{-}}) \hspace*{-0.5mm}=  \hspace*{-0.5mm}\exp \hspace*{-1mm} \left (
    \mathop{{}\E}_{\x \sim \pi({{C^{-}_{\x}}})} \hspace*{-1.0mm}
    \Big [\hspace*{-0.5mm} -\hspace*{-0.5mm} \gamma \;c \big(\hspace*{-0.5mm}\x, \phi(\x, \Wcal) ; \Ccal^{-}\hspace*{-0.5mm}\big) \hspace*{-0.5mm}\Big ]
    \right )\hspace*{-1mm}
\end{align}

In (\ref{eq:negative_prior}), $c(\x, \y; \Ccal^{-})$ is a classifier function that encodes softly whether or not $(\x, \y)$ is in $\Ccal^{-}$, which allows black-box use with any inference technique:
\begin{align}\label{eq:negative_constraint_function}
    c(\x, \y; \Ccal^{-}) = \sum_{j=1}^J ~ \prod_{k=1}^{K_j} ~ \sigma_{\tau_0, \tau_1} \Big ( f_{j} \big( \x, \y \big)_k \Big )
\end{align}
The definition of $c(\x, \y; \Ccal^{-})$ assumes that the negative region $\Ccal^{-}$ is defined by $J$ sets of $K_j$ inequality constraints $f_j(\x, \y) \leq 0$, i.e. $\Ccal^{-} = \bigcup_{j=1}^J \Ccal^{-}_{j}$ with $\Ccal^{-}_j = \{(\x, \y) ~|~ f_j(\x, \y) \leq 0\}$,
which can define arbitrary linear and nonlinear shapes in the input-output space.
$\sigma_{\tau_0, \tau_1}(z)$ is a soft indicator of whether a constraint of the form $z \leq 0$ is satisfied, a more generally-parameterizable sigmoidal activation defined as
\begin{align}\label{eq:negative_constraint_classifier}
    \hspace*{-2mm}\sigma_{\tau_0, \tau_1}(z) = \big( \tanh (-\tau_0 z) + 1 \big) \big( \tanh (-\tau_1 z) + 1 \big)
\end{align}
If \emph{all} constraints for at least one infeasible region $\Ccal^{-}_j$ are satisfied, our prior knowledge is violated and $c(\x, \y; \Ccal^{-})$ is far from 0.
Otherwise, at least one constraint of all infeasible regions is violated and our prior beliefs satisfied; $c(\x, \y; \Ccal^{-})$ is close to 0.
Contrary to other classification functions, the product of two tanh functions with different scales $\tau_0, \tau_1$ enables a sharp and steep overall classification of violating values in $z > 0$ and a smoother and flatter classification for satisfying values in $z \leq 0$, making gradients less vanishing for constraint-satisfying, i.e. region-violating inputs. We use $\tau_0 = 15, \tau_1 = 2$.

\section{Inference}
\label{app:inference}

Constraint priors can be substituted for the traditional prior term $p(\Wcal)$ with any black-box sampling or variational inference (VI) algorithm.
Here, we provide a summary of the algorithms we use and describe the trivial modifications used to incorporate constraint priors $p_{\Ccal}(\Wcal)$.
Note that the general form of $p_{\Ccal}(\Wcal)$ is not normalized, which does not pose a problem for inference in practice.

\textbf{Hamiltonian Monte Carlo (HMC)} HMC \cite{hmc} is a MCMC method  considered to be the ``gold standard'' in posterior sampling even though not being scalable. We substitute $p(\Wcal)$ by $p_{\Ccal}(\Wcal)$ in the potential energy term $U(\Wcal)$ computed at each sampling iteration:
\begin{equation}
    U(\Wcal) = - \log p_{\Ccal}(\Wcal) - \log p(\Dcal|\Wcal)
\label{eq:key_hmc}
\end{equation}
As the presence of $g(\Wcal|\Ccal)$ increases the magnitude of the prior $p_{\Ccal}(\Wcal)$, empirical performance typically improves by using a smaller step-size than with $p(\Wcal)$ for the same dataset.

\textbf{Stein Variational Gradient Descent (SVGD)} SVGD \cite{svgd} is a VI method where a set of $S$ particles (in our case, $\{\Wcal_s\}^S_{s=1}$) are optimized via functional gradient descent to mimic the true posterior. SVGD combines the efficiency of VI methods with the ability of MCMC methods to capture more expressive posterior approximations. $p(W)$ is substituted by $p_{\Ccal}(\Wcal)$ in the computation of the functional gradient:
\begin{align}
\hat{\phi}^*(\Wcal) = \frac{1}{S}\sum^{S}_{s=1}\Big[ &k(\Wcal_s, \Wcal) \nabla_{\Wcal_s}[\log p_{\Ccal}(\Wcal_s)\\
&+ \log p(\Dcal|\Wcal_s)] + \nabla_{\Wcal_s}k(\Wcal_s, \Wcal) \Big]\nonumber
\label{eq:key_svgd}
\end{align}
Our implementation of SVGD uses the weighted RBF kernel $k(x,x') = \exp ( - \tfrac{1}{h} || x - x'||_2^2)$ and adapting bandwith $h$ as suggested in \cite{svgd} as well as mini-batched data $\mathcal{D}$ for tractability.



\section{Experimental Details}
\label{app:experimental}

\subsection{Synthetic Examples}

For all experiments, the BNN used comprises a single hidden layer with 10 nodes, and Radial Basis Function (RBF) activations $\sigma(x) = \exp\{-x^2\}$.

All regression plots show the posterior mean function (bold line) as well as the confidence intervals for $\sigma=1$ (dark shading) and $\sigma=2$ (light shading).

\textbf{Figure \ref{fig:ex14}:} \textbf{(left)} The constrained regions are $y<2.5$ and $y > 3$ for $x \in [-0.3, 0.3]$. The function generating the training points is $y=-x^4+3x^2+1$. The negative prior formulation is used. \textbf{(right)} The input space is 2-dimensional and there are 3 classes (color-coded) with 8 training points in each class, generated from the Gaussian means $(-3, 1), (0, -3)$ and $(2, 3)$. The constrained region is $[1, 3] \times [-2, 0]$ and defined such that points within the box \textit{should} be classified as green. The positive prior is used. HMC (10000 burn-in, 1000 samples collected at intervals of 10) is used for both examples.

\textbf{Figure \ref{fig:ex35}:} \textbf{(left)} The positive constraints are $y =-x+5$ for $x \in [-5.0, -3.0]$ and $y=x+5$ for $x \in [3.0, 5.0]$. Both constraints are Gaussian with the $\sigma = 0.5$. The 3 training points are arbitrarily defined. HMC (10000 burn-in, 1000 samples collected at intervals of 10) is used. \textbf{(Right)} The constrained boxed region is $x \in [-1.0, 1.0]$ and $y \in [-5.0, 3.0]$. The function generating the training points is $y=-x^4+3x^2+1$. SVGD with 75 particles is used with Adagrad.

\subsection{Clinical action prediction}

For all experiments, the BNN used comprises a 2 hidden layers of 200 nodes each and RBF activations. SVGD is used for inference with 50 particles, 1500 iterations, Adagrad optimization, and a suitable batch size. The size of the full dataset is 298K; this reduces to 125K when points in the constrained region are filtered out. Details on the prior formulation for can be found in \ref{app:constraints}. The Dirichlet parameter is set to 10 for allowed classes and 0.01 for forbidden classes.

\subsection{Human motion prediction}
For these experiments, the BNN used comprises a 2 hidden layers of 100 nodes each and RBF activations. For inference, we again used SVGD and Adagrad with 50 particles and 1000 iterations. The negative prior used $50$ samples from $\pi(\Ccal^{-}_{\x})$ and $\gamma = 10,000$, see Eq. \ref{eq:negative_prior}.

We randomly chose a subset of 10 right-handed reaching trajectories from \cite{kratzer2019data}. This data was randomly split into 5 training and 5 test trajectories, which amounts to 243 train Markov states of sensors for training and 142 states for evaluation. Given this problem setting, the regression task had 12-dimensional inputs and 4-dimensional targets. The number of training trajectories was kept low to increase sparsity and the difficulty of successful robust generalization.

\section{Additional Results}
\label{app:additional}

\subsection{Additional Synthetic Examples}

\begin{figure}[H]
\centering
\includegraphics[width=0.48\columnwidth]{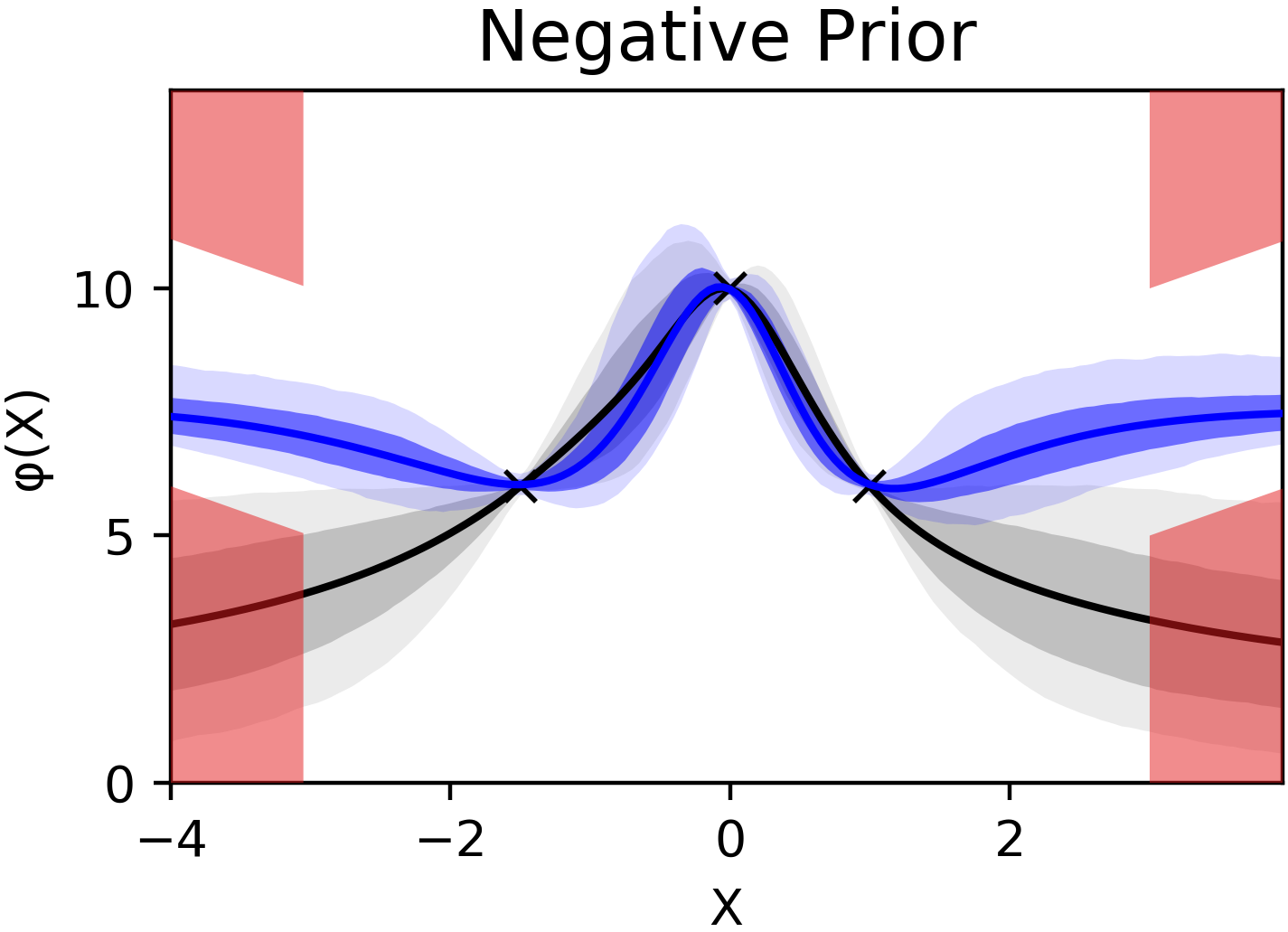}
\includegraphics[width=0.50\columnwidth]{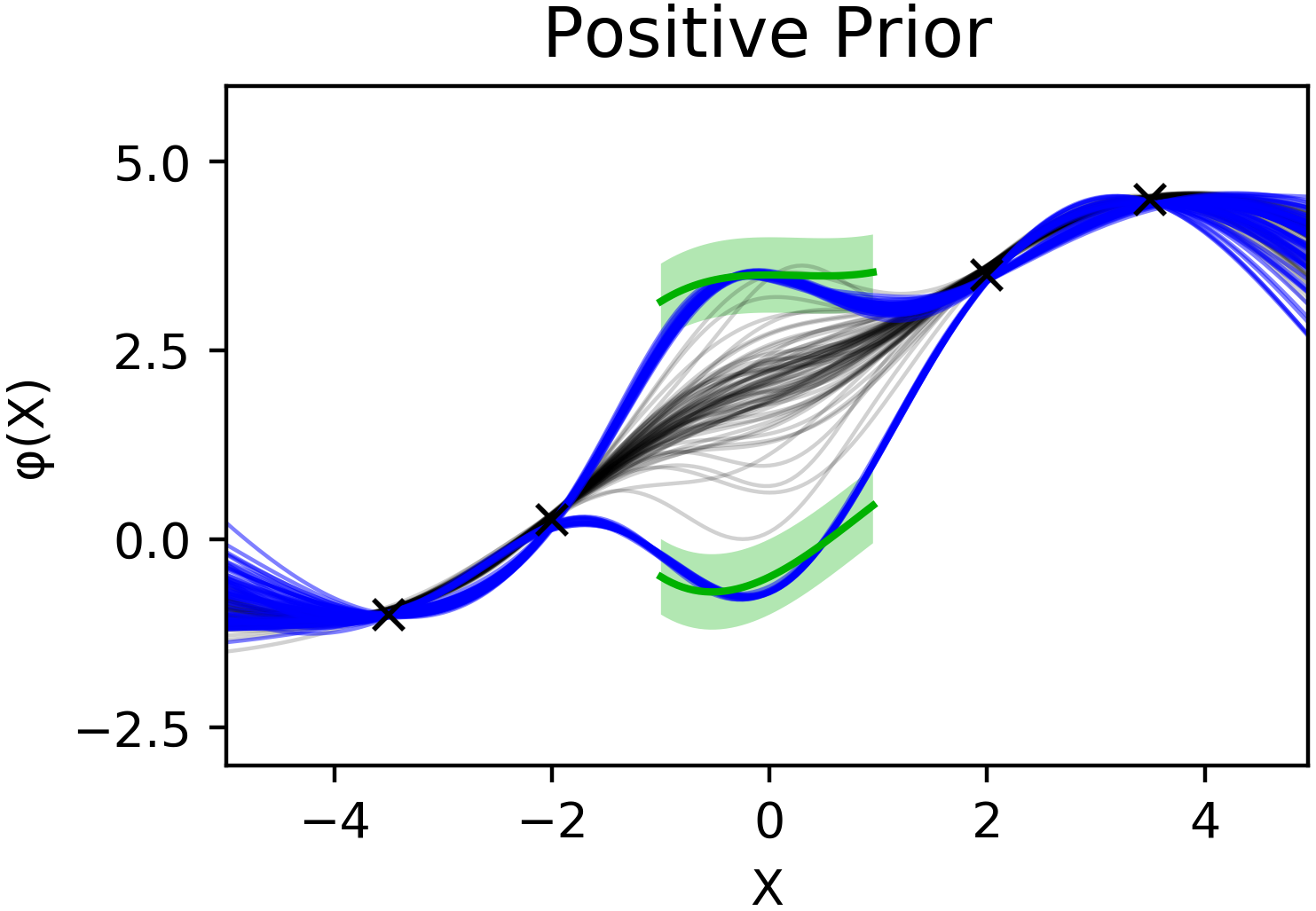}
\caption{\textbf{(left)} The same training set as Figure~\ref{fig:ex35} (left), but with negative constraints defined out-of-distribution. OC-BNNs fit the sparse data while avoiding the constraints. \textbf{(right)} Positive prior with mixture of two Gaussians. Using SVGD, individual OC-BNN samples (blue) capture both modes.}
\label{fig:ex26}
\end{figure}

Figure~\ref{fig:ex26} shows additional examples for out-of-distribution and multimodal behavior. \textbf{(left)}  Out-of-distribution negative constraints. The negative constraints are $y > -x+7$ and $y < -x+2$ for $x \in [-5.0, -3.0]$ and $y > x+7$ and $y <x+2$ for $x \in [3.0, 5.0]$. The training points are identical to those in the left plot of Figure~\ref{fig:ex35}. HMC (10000 burn-in, 1000 samples collected at intervals of 10) is used. \textbf{(right)} Multimodal positive constraints. The two positive functions are $y = -0.2x^3 + 0.5x^2 + 0.7x - 0.5$ and $y=0.2x^3 - 0.15x^2 + 3.5$, both for the domain $x \in [-1.0, 1.0]$. The training points were arbitrarily defined. An equally-weighted mixture of two Gaussians with $\sigma = 0.5$ is used as the positive constraint prior. SVGD with 75 particles and Adagrad are used.

\begin{thebibliography}{11}
\providecommand{\natexlab}[1]{#1}
\providecommand{\url}[1]{\texttt{#1}}
\expandafter\ifx\csname urlstyle\endcsname\relax
  \providecommand{\doi}[1]{doi: #1}\else
  \providecommand{\doi}{doi: \begingroup \urlstyle{rm}\Url}\fi

\bibitem[Hafner et~al.(2018)Hafner, Tran, Lillicrap, Irpan, and Davidson]{ncp}
Hafner, D., Tran, D., Lillicrap, T., Irpan, A., and Davidson, J.
\newblock Reliable uncertainty estimates in deep neural networks using noise
  contrastive priors.
\newblock In \emph{eprint arXiv:1807.09289}, 2018.

\bibitem[Johnson et~al.(2016)Johnson, Pollard, Shen, Li-wei, Feng, Ghassemi,
  Moody, Szolovits, Celi, and Mark]{mimic}
Johnson, A.~E., Pollard, T.~J., Shen, L., Li-wei, H.~L., Feng, M., Ghassemi,
  M., Moody, B., Szolovits, P., Celi, L.~A., and Mark, R.~G.
\newblock Mimic-iii, a freely accessible critical care database.
\newblock \emph{Scientific data}, 3:\penalty0 160035, 2016.

\bibitem[Kratzer(2019)]{kratzer2019data}
Kratzer, P.
\newblock mocap-mlr-datasets.
\newblock \url{https://github.com/charlespwd/project-title}, 2019.

\bibitem[Kratzer et~al.(2018)Kratzer, Toussaint, and
  Mainprice]{kratzer2018towards}
Kratzer, P., Toussaint, M., and Mainprice, J.
\newblock Towards combining motion optimization and data driven dynamical
  models for human motion prediction.
\newblock In \emph{2018 IEEE-RAS 18th International Conference on Humanoid
  Robots (Humanoids)}, pp.\  202--208. IEEE, 2018.

\bibitem[Liu \& Wang(2016)Liu and Wang]{svgd}
Liu, Q. and Wang, D.
\newblock Stein variational gradient descent: A general purpose bayesian
  inference algorithm.
\newblock In \emph{Advances In Neural Information Processing Systems}, pp.\
  2378--2386, 2016.

\bibitem[Lorenzi \& Filippone(2018)Lorenzi and Filippone]{lorenzi}
Lorenzi, M. and Filippone, M.
\newblock Constraining the dynamics of deep probabilistic models.
\newblock \emph{arXiv preprint arXiv:1802.05680}, 2018.

\bibitem[MacKay(1995)]{mackay}
MacKay, D. J.~C.
\newblock Probable networks and plausible predictions -- a review of practical
  bayesian methods for supervised neural networks.
\newblock In \emph{Network: Computation in Neural Systems, 6:3, 469-505}, 1995.

\bibitem[Namdari et~al.(2012)Namdari, Yagnik, Ebaugh, Nagda, Ramsey,
  Williams~Jr, and Mehta]{namdari2012defining}
Namdari, S., Yagnik, G., Ebaugh, D.~D., Nagda, S., Ramsey, M.~L., Williams~Jr,
  G.~R., and Mehta, S.
\newblock Defining functional shoulder range of motion for activities of daily
  living.
\newblock \emph{Journal of shoulder and elbow surgery}, 21\penalty0
  (9):\penalty0 1177--1183, 2012.

\bibitem[Neal(1995)]{neal}
Neal, R.~M.
\newblock \emph{Bayesian Learning for Neural Networks}.
\newblock PhD thesis, Graduate Department of Computer Science, University of
  Toronto, 1995.

\bibitem[Neal(2012)]{hmc}
Neal, R.~M.
\newblock Mcmc using hamiltonian dynamics.
\newblock In \emph{Handbook of Markov Chain Monte Carlo}, 2012.

\bibitem[Sun et~al.(2019)Sun, Zhang, Shi, and Grosse]{fbnn}
Sun, S., Zhang, G., Shi, J., and Grosse, R.
\newblock Functional variational bayesian neural networks.
\newblock \emph{arXiv preprint arXiv:1903.05779}, 2019.

\end{thebibliography}
\end{document}